# Automatic pain recognition from Blood Volume Pulse (BVP) signal using machine learning techniques

Fatemeh Pouromran[1], Yingzi Lin[1], and Sagar Kamarthi[1,*]

[1] Department of Mechanical and Industrial Engineering, Northeastern University, Boston, Massachusetts, United States of America.

[*] Corresponding author: sagar@coe.neu.edu

## Abstract

Physiological responses to pain have received increasing attention among researchers for developing an automated pain recognition sensing system. Though less explored, Blood Volume Pulse (BVP) is one of the candidate physiological measures that could help objective pain assessment. In this study, we applied machine learning techniques on BVP signals to device a non-invasive modality for pain sensing. Thirty-two healthy subjects participated in this study. First, we investigated a novel set of time-domain, frequency-domain and nonlinear dynamics features that could potentially be sensitive to pain. These include 24 features from BVP signals and 20 additional features from Inter-beat Intervals (IBIs) derived from the same BVP signals. Utilizing these features, we built machine learning models for detecting the presence of pain and its intensity. We explored different machine learning models, including Logistic Regression, Random Forest, Support Vector Machines, Adaptive Boosting (AdaBoost) and Extreme Gradient Boosting (XGBoost). Among them, we found that the XGBoost offered the best model performance for both pain classification and pain intensity estimation tasks. The ROC-AUC of the XGBoost model to detect low pain, medium pain and high pain with no pain as the baseline were 80.06 %, 85.81 %, and 90.05 % respectively. Moreover, the XGboost classifier distinguished medium pain from high pain with ROC-AUC of 91%. For the multiclass classification among three pain levels, the XGBoost offered the best performance with an average F1-score of 80.03%. Our results suggest that BVP signal together with machine learning algorithms is a promising physiological measurement for automated pain assessment. This work will have a national impact on accurate pain assessment, effective pain management, reducing drug-seeking behavior among patients, and addressing national opioid crisis.

# Introduction

Pain is an unpleasant sensory and emotional experience as well as a primary symptom of many medical conditions. Effective pain management is one of the main goals in patient healthcare. Therefore, an accurate pain assessment is necessary to diagnose and provide a proper treatment plan. However, since there is no automated pain estimation method, clinicians rely on patient's self-report about how much pain they are experiencing. The most common available measures for pain assessment are numerical rating scales (NRS), and verbal rating scales (VRS)[1]. Unfortunately, these self-reporting measures are limited by the fact that they require the patient to be functionally capable, mentally alert, and clinically cooperative. For example, self-reporting is not a feasible option for patients with dementia or paralysis or patients who are drowsy or unable to answer consciously correct. Moreover, the absence of proper objective pain assessment tools to quantify the pain severity has resulted in sub-optimal treatment plans, delayed responses to patient needs, over-prescription of opioids, and drug-seeking behavior among patients.

According to the biological mechanism of pain, multiple areas of the Autonomic Nervous System (ANS) are participated in experiencing the pain. Pain often starts with the activation of the sensory neural pathway upon stimulation by noxious mechanical, heat, cold, chemical, or inflammatory stimuli. ANS has two parts: (1) Parasynaptic Nervous System (PSN) activated during rest, and (2) Sympathetic Nervous System (SNS) activated during stress or pain. Activation of ANS leads to changes in the electrical property of the brain, heart, muscle, and skin. These changes can be measured by physiological signals such as Electroencephalography (EEG), Electrocardiography (ECG), Blood Volume Pulse (BVP), and Electrodermal Activity (EDA)[2–7]. So, these signals are potential candidates for automatically detecting the existence of pain and estimating its intensity. Chen et al.[8], have recently reviewed the mechanism of pain and various wearable physiological and behavioral sensors used in the healthcare domain that may be helpful for automated monitoring systems for pain and stress detection.

Developing an automated pain assessment system has received increasing attention among researchers. Several researchers have studied automatic pain assessment using machine learning techniques. However, there are only a few research groups that collected databases of physiological signals specific to pain. Walter et al. [9] introduced the BioVid heat pain database, in which they induced four levels of gradually increasing pain through temperature elevation. They recorded video streams, EDA, ECG, and EMG signals from 87 healthy subjects during the experiment. The pain labels for the recorded data were based on the four levels of temperature for pain elicitation. Researchers have studied physiological signals from the BioVid database to build machine learning models for pain detection or pain intensity estimation tasks[5,10–14]. The results shows that EDA signal works significantly better than EMG and ECG signal for automated pain intensity estimation [5].

Aung et al.[15] proposed the multimodal EmoPain Dataset for automatic detection of chronic pain-related expressions. This dataset consists of 22 individuals suffering from chronic lower back pain and 28 health subjects carrying out physical exercises. They recorded face videos, head-mounted and room audio signals, full-body 3D motion capture, and EMG signals from back muscles. Two sets of labels were assigned: First, the level of pain from facial expressions was annotated by eight raters who gave values between 0 and 1. Second, the occurrence of six pain-related body behaviors (guarding or stiffness, hesitation, bracing or support, abrupt action, limping, rubbing, or stimulating) was segmented by four experts. Most recently, Velana et al.[16] introduced SenseEmotion Database, consisting of video streams, trapezius EMG, respiratory (RSP), ECG, and EDA. The experiment was conducted on 45 healthy participants, each subjected to a series of artificially induced heat pain stimuli. The heat pain was induced at three levels depending on temperature. These temperatures were separately calculated for each participant. Thiam et al.[17] studied the modalities in this dataset for the recognition of artificially induced heat pain. Pouromran et al. [18] explored EDA signal collected through Cold Pressor Test from 29 healthy participants and built four-category pain intensity classification model using deep learning generated representations of the signal.

Gruss et al.[19] proposed a psychophysiological experiment in which 134 healthy participants were subjected to thermal and electrical pain stimuli, while audio, video, and physiological signals such as EMG, ECG, and Skin Conductance Level were recorded as X-ITE Pain Database. These datasets can be categorized by the source of pain, recorded modality, ground truth, and subject groups. The subject groups can be categorized by age group, for example, infants [20], and health condition of the subjects. A better overview of the automatic pain assessment studies can be found in [7,21,22].

Blood volume pulse signal reflects the changes in blood volume in tissue in each beat (pulse) of the heart. It is measured by shining an infrared light through the body's surface, usually a finger, then the amount of infrared light that returns to the sensor is recorded. The light registered by the sensor is proportional to the amount of blood volume in a tissue. BVP is widely used to measure heart rate variability (HRV), which consists of changes in the time intervals between consecutive heartbeats called inter-beat intervals (IBIs) [23]. In comparison with the ECG sensor and Respiration sensor which are also used for HRV measurements, the BVP sensor is easier to apply and preferred in some clinical situations, for example, when a subject is not still and stationary. BVP is considered as a candidate sensor to reflect the pain response in the autonomic nervous system that affects heart activity. However, the BVP signal is hardly unexplored in the literature for automated pain assessment. To our knowledge, chu et al.[24] are the only research group that investigated BVP in the pain experiment. They collected BVP, ECG, and EDA signals from six healthy subjects aged between 22 and 25 years old to classify the level of pain induced by electrical stimulation. They extracted simple statistical features, including mean, standard deviation, minimum, maximum, range, minimum ratio, and maximum ratio from the signal and the first-order differences of the signal. In their paper, they used 16 statistical features from the time-domain of BVP, plus eight features from ECG and ten features from EDA to predict pain levels using Linear Discriminant Analysis and principal component analysis. Chu et al.[25] also developed a hybrid genetic algorithm using support vector machines (SVMs) and k-nearest neighbors (KNN) to detect different pain levels on the same dataset.

In this study, we aim to investigate the BVP signal for automated pain intensity classification. Toward this objective, we first filtered the BVP signal and extracted a broad set of 44 features from time-domain, frequency-domain, and nonlinear measures, directly from the BVP and corresponding Inter Beat Intervals (IBIs) captured from BVP. Then we utilized machine learning algorithms to build an automatic pain assessment model and find the best pain-sensitive features from the BVP signal. These features captured the sympathetic activation changes that occur in response to cold pain. These physiological responses were acquired from a finger-clip BVP sensor during the cold pain experiment. We also explored the feature importances and conducted statistical analysis to test the potential relationship between features and pain states.

## Materials and Methods

### Data Collection from Cold Pressor Test

The data was collected through the cold pain experiment at the Intelligent Human-Machine Systems lab at Northeastern University (IRB#: 191215) to explore human physiological signals affected by pain. We obtained informed consent from all the participants involved for participation in the experiments. All methods were carried out in accordance with relevant guidelines and regulations. The participants were informed that they could stop this experiment at any time. A total of thirty-two healthy subjects (6 females and 26 males) aged 18 to 24 participated in this experiment. Each subject was asked to rest when a 20-second recording of the Blood Volume Pulse (BVP) sensor was taken as the baseline measurement. Simultaneously, the subject was asked to focus on a green dot displayed on a monitor in front of them. After initial baseline 20 seconds, the subject was asked to place their dominant hand into a bucket containing iced water to trigger pain from cold temperature (32F or 0C). Then, the BVP signal of the subject was recorded for the 200s. The sensor was placed on the subject's middle finger for recording BVP. Every 20 seconds, the subjects were asked to self-report their pain score based on a numerical rating scale of 0 to 10 provided to them during the experiment. For each subject, we recorded a total of 220 seconds of BVP signal at a fixed sampling rate of 2048 Hz. Figure 1 shows the average and standard deviation of these self-reported pain intensities during the cold pressor test.

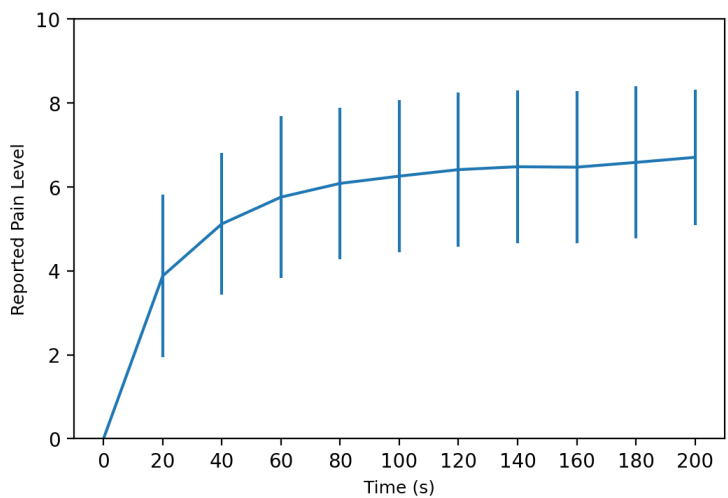

**Figure 1**. Mean and standard deviation of pain levels reported by 32 healthy subjects during the cold pressor test.

**BVP Signal Preprocessing and Inter Beat Intervals Detection**

Before extracting feature representations from a physiological signal, it is necessary to undertake a preprocessing step to improve the signal-to-noise ratio (SNR). We employed an 8-Hz Butterworth low-pass filter to reduce the noise and artifacts within the BVP signal.

Since the BVP signal measures blood volume changes in vessels due to heart activity, the peaks in BVP are associated with heartbeats. So, we used BVP to identify the Inter Beat Intervals (IBI), the distances between consecutive heartbeats, also known as RR intervals. Figure 2 presents a 40-second segment of BVP and corresponding IBI signal. We investigated changes in IBIs to study the Heart rate variability (HRV), which indexes neurocardiac function and is generated by heart-brain interactions and dynamic nonlinear Autonomic Nervous System (ANS) processes[23].

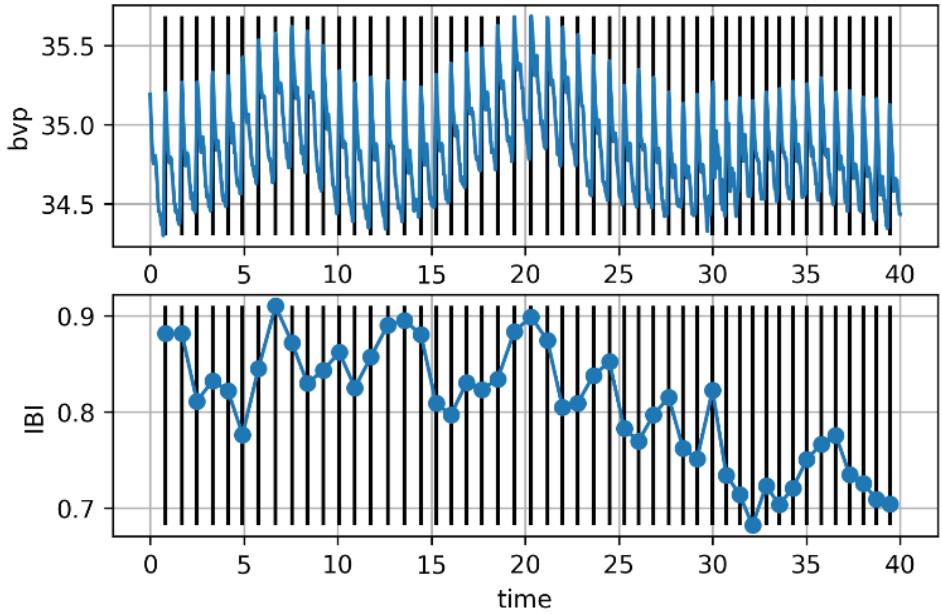

**Figure 2.** Sample recording of BVP signal and its corresponding IBIs extracted from BVP for a subject before and after the experiment. The first 20 seconds is the baseline in which there is no pain, and the following 20 seconds is when the subject is experiencing pain.

**Feature Engineering**

We extracted features from time-domain, frequency-domain, and nonlinear metrics from BVP and its corresponding Inter-beat Intervals (IBIs). After noise removal, the data was segmented into 5-second sliding windows with fifty percent overlap. We extracted 24 features directly from BVP signal and 20 features from IBIs related to the heart rate variability measured by BVP signal. Lastly, the measurements were normalized for each individual subject by subtracting the mean and dividing by the standard deviation. The feature vector is generated by concatenating the set of features extracted from IBIs with those extracted directly from BVP, resulting in a feature vector of dimensionality 20 + 24 = 44. The IBIs features were selected from the heart rate variablity metrics widely used in other domians[23]. These metrics and their description are listed in Table 1. The BVP features were selected from the time-series features which were reported to be the most effective in a wide variety of other applications[26]. Table 2 shows the list of these features and their description. In authors' earlier work, they found these features to be effective for heat pain assessment using EDA, ECG and EMG signals[5]. We performed the feature extractions using the Scipy[27], pyphysio[28], catch22[26], Pandas[29] and Numpy[30] libraries in Python. To the best of our knowledge, this is the first study that investigated this novel set of features from BVP and its corresponding IBIs for objective pain assessment.

| Number | Symbol | Feature Description |
|---|---|---|
| 1 | RMSSD | Root mean square of successive time differences between heartbeats |
| 2 | SDSD | Standard deviation of the 1st order discrete differences |
| 3 | pNN50 | Percentage of successive IBIs that differ by more than 50 milliseconds |
| 4 | pNN25 | Percentage of successive IBIs that differ by more than 25 milliseconds |
| 5 | pNN10 | Percentage of successive IBIs that differ by more than 10 milliseconds |
| 6 | RRmean | Mean of RR intervals |
| 7 | RRSTD | Standard deviation of RR intervals |
| 8 | RRMed | Median of RR intervals |
| 9 | RRMin | Minimum of RR intervals |
| 10 | RRMax | Maximum of RR intervals |
| 11 | I-VLF | Power in very low frequency [0.003, 0.04 Hz] |
| 12 | I-ILF | Power in low-frequency band [0.04, 0.15 Hz] |
| 13 | I-HF | Power in high-frequency band [0.15, 0.4 Hz] |
| 14 | I-Pow | Total power of IBIs |
| 15 | I-SD1 | Poincaré plot standard deviation perpendicular the line of identity |
| 16 | I-SD2 | Poincaré plot standard deviation along the line of identity |
| 17 | I-SD12 | Ratio of SD1-to-SD2 |
| 18 | I-Sdell | SD1*SD2*pi value of the Poincare' plot of input IBIs |
| 19 | I-DFA1 | Detrended fluctuation analysis, which describes short-term fluctuations of IBIs |
| 20 | I-ApEn | Approximate entropy which measures the regularity and complexity of IBIs |

**Table 1.** Feature engineering from Inter-beat Intervals (IBIs) based on BVP signal.

| Number | Symbol | Feature Description |
|---|---|---|
| 1 | Mean | Mean of BVP signal |
| 2 | STD | The standard deviation of BVP signal |
| 3 | DNH5 | Mode of 5-bin z-scored histogram |
| 4 | DNH10 | Mode of 10-bin z-scored histogram |
| 5 | COf1eac | First 1/e crossing of the autocorrelation function |
| 6 | AMI | Auto-mutual information, m = 2, τ = 5 |
| 7 | DNn | Time intervals between successive extreme events below the mean |
| 8 | COf1eac | First 1/e crossing of the autocorrelation function |
| 9 | COFmin | First minimum of the autocorrelation function |

| 10 | SPow5th | Total power in the lowest fifth of frequencies in the Fourier power spectrum |
|---|---|---|
| 11 | SPowCent | Centroid of the Fourier power spectrum |
| 12 | FCmean | Mean error from a rolling 3-sample mean forecasting |
| 13 | COtrev | Time-reversibility statistic, $((x_{t+1} - x_t)3)t$ |
| 14 | AMI | Auto-mutual information, $m = 2$, $\tau = 5$ |
| 15 | INAut | First minimum of the auto-mutual information function |
| 16 | MDpnn40 | Proportion of successive differences exceeding $0.04\sigma$ |
| 17 | SBlongst | Longest period of successive incremental decreases |
| 18 | SBshanEn | Shannon entropy of two successive 3-letter symbolization |
| 19 | SBTrace | Trace of covariance of transition matrix between symbols in 3-letter alphabet |
| 20 | SBPerioc | Periodicity measure of Wang |
| 21 | FCmean | Change in correlation length after iterative differencing |
| 22 | COexpfit | Exponential fit to successive distances in 2-d embedding space |
| 23 | SCFlucdfa | Proportion of slower timescale fluctuations that scale with DFA |
| 24 | SCFlucrsr | Proportion of slower timescale fluctuations that scale with linearly rescaled range fits |

**Table 2.** Feature engineering directly from Blood Volm Pulse (BVP) signal

**Model Architecture for Pain Assessment**

We categorized the 0 to 10 self-reported pain intensities into four pain states as no pain (NP: $P = 0$), low pain (LP: $0 < P \leq 3$), medium pain (MP: $3 < P \leq 6$), and high pain (HP: $6 < P \leq 10$). Then we performed a set of binary classification tasks between pair of these pain states. We also trained models for a multiclass classification task to differentiate the pain levels during the cold pressor test. In addition to classification, we performed the regression to predict the continuous pain intensity as the pain levels are ordinal.

We explored different machine learning algorithms in this study: Logistic Regression (LR), Support Vector Machines (SVM), and Ensemble methods including Random Forest (RF), Adaptive Boosting (AdaBoost) and Extreme Gradient Boosting (XGBoost). The ensemble method is an algorithm that uses a group of predictors, called Ensembles, generally yielding an overall better model. Random Forest is an ensemble of Decision Trees, generally trained via the bagging method, or sometimes pasting. When sampling is performed with replacement, this method is called *bagging*, which is short for bootstrap aggregating. When sampling is performed without replacement, it is called *pasting*. In other words, both bagging and pasting allow training instances to be sampled several times across multiple predictors, but only bagging allows training instances to be sampled several times for the same predictor. One way to get a diverse set of classifiers is to use the same training algorithm for every model and train them on different random subsets of the training set. Once all models are trained, the ensemble can make a prediction for a new instance by simply aggregating the predictions of all models. Generally, the ensemble has a similar bias but a lower variance than a single model trained on the original training set. The random forest algorithm also introduces extra randomness when growing trees; it searches for the best feature among a random subset of features, instead of searching among all features, when splitting a node. The algorithm results in greater tree diversity, which trades a higher bias for a lower variance to generally obtain a better model.

Boosting is another ensemble method. The general idea of most boosting methods is to train models sequentially, each trying to correct its predecessor. The most popular boosting methods are Adaptive Boosting (AdaBoost) and Gradient Boosting[31]. In AdaBoost, a new model corrects its predecessor by paying more attention to the training instances that the predecessor under-fitted. This results in new models focusing more and more on the hard cases. In Gradient Boosting, instead of tweaking the instance weights at every iteration as AdaBoost does, the method tries to fit the new predictor to the residual errors made by the previous predictor. XGBoost is an advanced implementation of a gradient boosting algorithm that is extremely fast, scalable, and portable.

We employed 5-fold stratified cross-validation among all subjects to evaluate our models. This cross-validation object is a variation of k-fold that returns stratified folds made by preserving the percentage of samples for each class. So, we generate test sets containing the same distribution of classes, as close as possible, which is a helpful tip to build a model on an imbalanced dataset when the classes are not represented equally. We also applied the ExtraTree-based feature selection technique to find the optimal subset of features for each model. This method computes the feature importance based on impurity measure in the forest of extremely randomized decision trees. We used 16% of the samples outside of the train set and test set for hyperparameter tuning and feature selection. We conducted the exhaustive grid search to find the balance between bias and variance and thus, prevent the model from underfitting and overfitting. We employed the L2 regularization technique, which adds squared magnitude of coefficient as a penalty term to the loss function to reduce the model complexity. We used Scikit-Learn and XGBoost in Python to build and tune the models.

**Handling the Class Imbalance**

To encounter the issue of class imbalance in our dataset, we used the Synthetic Minority Oversampling Technique (SMOTE) [32,33]. In this technique, we oversample the minority class by creating synthetic examples rather than by resampling with replacement. SMOTE first selects a minority class instance at random and finds its *k* nearest minority class neighbors. Then randomly chooses one of these neighbors and connects them to form a line segment in the feature space. The synthetic instances are generated as a convex combination of the two chosen instances. Using this data augmentation technique, we generated as many synthetic examples for the minority class as required to balance the class distribution in the training dataset.

**Model Evaluation Metrics**

The model performance was measured by Accuracy, Area under the ROC curve (ROC-AUC), Balanced Accuracy (B-ACC), Precision, Recall, and F1-score. It is noted that F1-score and ROC AUC provide a less biased estimate of the performance when the classes are imbalanced. The below equations show the mathematical expression of precision, recall, and F1-score, where TP, FP, TN, and FN refer to "True Positive," "False Positive," "True Negative,", and "False Negative", respectively. Since our data is imbalanced, we also calculate the balanced accuracy (B_Acc), where $m$ is the number of classes in our dataset.

$$Precision = \frac{TP_i}{TP_i + FP_i}$$

$$Recall = \frac{TP_i}{TP_i + FN_i}$$

$$F1 = \frac{2.Precision.recall}{Precision + recall}$$

$$BalacnedAccuracy = \frac{1}{m}\sum_{i=1}^{m}\left(\frac{TP_i}{TP_i + FN_i}\right)$$

To evaluate the performance of regressor models, we considered Mean Absolute Errors (MAE) and Root Mean Square Errors (RMSE) as performance metrics. The equations below show the mathematical expression of these performance metrics: $y$ is the actual value, $\hat{y}$ is the predicted value, and n is the number of samples.

$$MAE = \frac{\sum_{k=1}^{n}|y-\hat{y}|}{n}$$

$$RMSE = \sqrt{\frac{\sum_{k=1}^{n}(y-\hat{y})^2}{n}}$$

**Statistical analysis between pain levels**

We evaluated the differences between pain levels for the different characteristics of BVP and IBIs during the cold pressor test. First, we tested the normality of features using the Kolmogorov-Smirnov test. Then, since the data were not Normal, we used Dunn's test to compare the values of the features at different pain levels. Dunn's test is a non-parametric multiple comparison test to pinpoint which specific groups are significantly different from the others after rejecting ANOVA null hypothesis. In this analysis, a $p$-value < 0.05 was considered statistically significant.

## Results and Discussion

### Overall Pain Classification Performances

We performed three binary classification tasks for pain detection. In each model, our goal was to detect the presence of pain. Table 3 presents the results of 5-fold stratified cross-validation for detecting the presence of pain (no pain vs. low pain, no pain vs. medium pain, no pain vs. high pain). The XGBoost model using the features extracted from the BVP signal resulted in pain detection with ROC-AUC of 90.05 ± 2.67% for high pain, 85.81 ± 5.06% for medium pain, and 80.06 ± 5.14% for low pain, with no pain as the baseline. The Receiver Operating Characteristics (ROC) is a curve of probability in which we plot the True Positive Rate (TPR) against the False Positive Rate (FPR). The area under the ROC curve measures the degree of separability of classes achieved by a classifier. It tells how much the model is capable of distinguishing between classes at different thresholds. The higher the AUC, the better the model is at distinguishing subjects with pain from those with no pain. For example, the average ROC-AUC of 90.05% means a 90.05% chance that the model can differentiate between no pain and high pain. Even if patients are in low pain, there is an 80.06% chance that the model can distinguish low pain from no pain condition. Please note that in the benchmark classifier whose accuracy depends only on the proportion of positive and negative cases in the dataset, the value for ROC-AUC would be 50%, which means that there is only a 50% chance to separate the classes accurately. The model's classification matrix is shown in Figure 3. It shows how samples in the presence and absence of pain can be classified if we set the threshold to 0.5 on the predicted class probabilities. This is an impressive model performance for pian detection using an easy-to-capture signal like BVP.

| Metric | No pain vs. Low pain | No pain vs. Medium pain | No pain vs. High pain |
|---|---|---|---|
| Accuracy (%) | 75.25 ± 2.77 | 85.0 ± 3.94 | 87.01 ± 1.41 |
| ROC-AUC (%) | 80.06 ± 5.14 | 85.81 ± 5.06 | 90.05 ± 2.67 |
| B-Acc (%) | 73.84 ± 3.02 | 75.56 ± 3.35 | 79.61 ± 3.52 |
| Precision (%) | 80.07 ± 3.12 | 92.04 ± 0.93 | 93.27 ± 1.47 |
| Recall (%) | 79.91 ± 4.25 | 89.72 ± 4.32 | 90.9 ± 1.01 |
| F1-score (%) | 79.89 ± 2.57 | 90.83 ± 2.59 | 92.06 ± 0.84 |

**Table 3.** Pain detection performance (mean ± std) of XGBoost model using BVP signal.

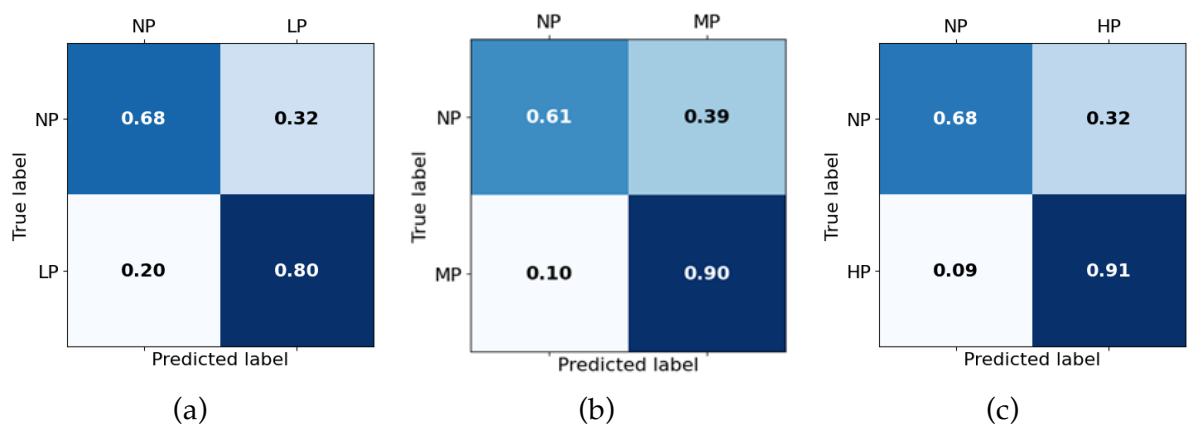

(a) (b) (c)

**Figure 3.** The classification matrix for detecting the presence of pain using BVP signal and XGBoost model during the cold pressor test: (a) no pain vs. low pain; (b) no pain vs. medium pain; (c) no pain vs. high pain. The pain becomes more detectable at higher levels of pain, as one can be expected.

We also performed three binary classifications to distinguish the different pairs of pain levels: Low pain vs. Medium pain, Low pain vs. High pain, and Medium pain vs. High pain.

Table 3 presents the result of 5-fold stratified cross-validation. The reported results in Table 4. represent that the BVP signal can distinguish between low and high pain with a ROC-AUC of 95.09 ± 0.91%. According to the results, there is an average 88.5 % chance to distinguish low pain from medium pain. There is also an average 83.41% chance to separate high pain from medium pain. The classification matrix of each classification task is shown in Figure 4. The highest F1-score is 95.09 % between low pain and high pain.

| Metric | Low vs. Medium | Low vs. High | Medium vs. High |
|:---:|:---:|:---:|:---:|
| Accuracy | 85.66 ± 0.97 | 89.97 ± 0.98 | 83.41 ± 1.49 |
| ROC-AUC | 88.5 ± 1.07 | 95.09 ± 0.91 | 91.31 ± 1.02 |
| B-ACC | 78.92 ± 1.64 | 85.96 ± 0.64 | 83.41 ± 1.49 |
| Precision | 89.33 ± 0.92 | 92.75 ± 0.35 | 82.84 ± 2.28 |
| Recall | 92.06 ± 1.21 | 93.97 ± 1.51 | 83.67 ± 3.04 |
| F1-score | 90.66 ± 0.63 | 93.35 ± 0.71 | 83.2 ± 1.57 |

**Table 4.** Performance of pain level binary classification (mean ± std) of XGBoost model using BVP signal.

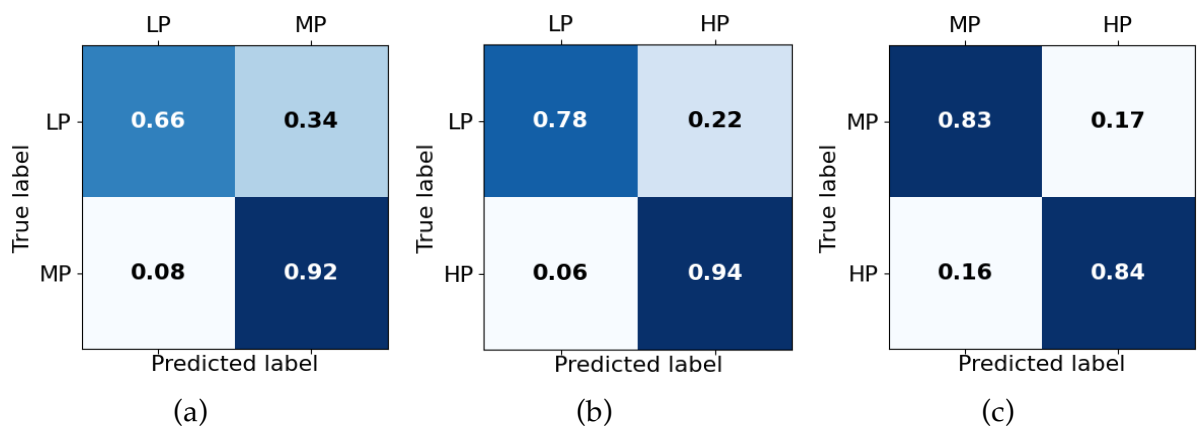

**Figure 4**. The classification matrix for classification of different levels of pain using XGBoost and BVP signal during the cold pressor test: (a) Low pain vs. Medium pain; (b) Low pain vs. High pain; (c) Medium pain vs. High pain.

**Multiclass Pain Intensity Classification and Regression**

We explored how the BVP signal identifies the pain intensity when we know that a patient is experiencing pain. We explored different machine learning models to classify pain into three classes: Low, Medium, and High pain. As shown in Table 5, we found that the XGBoost model gives the best results with an average F1-score of 80.03 ± 1.28%, balanced accuracy of 78.09 ± 1.39%, and ROC-AUC of 92.25 ± 0.95%. Figure 5 presents the classification matrics and ROC-AUC for XGBoost for the 3-class pain level classification. Table 6 summarizes classification performance of XGBoost model. This result is considerably better than the naïve benchmark rule in which all samples are classified as medium pain, which is the majority class in our dataset. As can be seen in the last row of Table 5, the balanced accuracy for the naive benchmark rule is 33% for the 3-class pain level classification, and the random average ROC-AUC value for that is 50%.

| Model | F1-score (%) | Balanced Accuracy (%) | ROC-AUC (%) |
|---|---|---|---|
| Logistic Regression | 44.01 ± 1.45 | 42.01 ± 2.02 | 59.39 ± 1.82 |
| Support Vector Machines | 53.33 ± 0.58 | 50.9 ± 1.56 | 68.33 ± 1.41 |
| Random Forest | 77.95 ± 1.86 | 75.02 ± 1.66 | 91.1 ± 0.77 |
| AdaBoost | 61.38 ± 2.05 | 59.62 ± 2.31 | 71.74 ± 1.18 |
| XGBoost | **80.03 ± 1.28** | **78.09 ± 1.19** | **92.25 ± 0.95** |
| Random benchmark | 26.58 ± 0.1 | 33.33 ± 0.0 | 50.0 ± 0.0 |

**Table 5.** Exploring model performance of different machine learning models for three-class pain intensity classification task using BVP signal during cold pressor test.

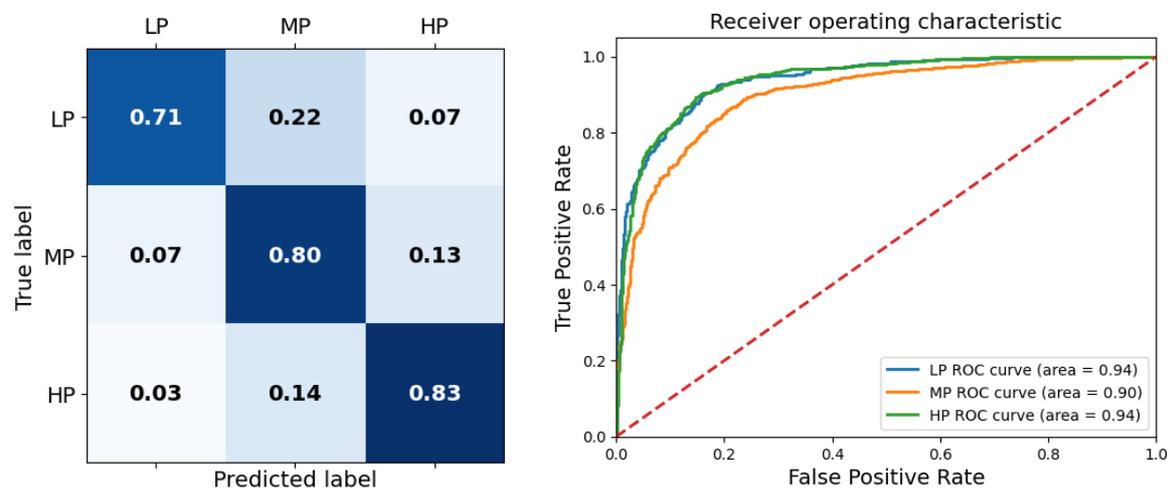

**Figure 5**. Classification matrix for pain intensity classification using XGBoost and BVP signals among three different levels of pain: low pain (LP), medium pain (MP), and high pain (SP).

| Pain level | Precision | Recall | F1-score |
|---|---|---|---|
| Low pain | 0.70 | 0.71 | 0.71 |
| Medium pain | 0.79 | 0.80 | 0.80 |
| High pain | 0.84 | 0.83 | 0.84 |

**Table 6.** Classification performance of XGBoost model for pain intensity classification using BVP during cold pressor test. The average accuracy is 80%.

Since the pain levels are ordinal, it is desirable to treat pain assessment as a regression task instead of a classification task. The cost of misclassifying high pain as a medium, or high pain as a low pain are not the same. A multiclass classification model treats both errors equally, while a regression model with continuous response values tends to minimize the distance between the predicted pain intensity and the actual pain intensity. For continuous pain estimation to predict pain intensity on a numerical rating scale between 1-10, we explored linear regression, support vector regressor, random forest, AdaBoost, and XGBoost. Performance evaluation of these models is presented in Table 7. XGboost gave the best pain estimation result with an average MAE of 0.94 ± 0.04 and an average RMSE of 1.26 ± 0.06 from 5-fold stratified cross-validation.

| Method | MAE | RMSE |
|---|---|---|
| Linear Regression | 1.63 ± 0.03 | 1.98 ± 0.04 |
| Support Vector Regressor | 1.57 ± 0.03 | 1.95 ± 0.03 |
| Random Forest | 1.06 ± 0.02 | 1.39 ± 0.04 |
| AdaBoost | 1.57 ± 0.03 | 1.88 ± 0.03 |
| XGBoost | **0.94 ± 0.04** | **1.26 ± 0.06** |
| Naïve Benchmark (P=5) | 1.83 ± 0.03 | 2.21 ± 0.03 |

**Table 7**. Pain intensity estimation performance of five different machine learning models using BVP during cold pressor test.

**Exploring Feature Importance for Pain Recognition**

To determine the contribution of each of the 44 features extracted directly from the raw BVP signals and indirectly from the signal IBIs, we calculated the impurity-based feature importance for each feature using 5-fold stratified cross-validation using ExtraTree classifier. Then we averaged the importance of features across the folds. Figure 6 presents the top features for pain classification with feature importance above the 0.025 threshold. See Table 1 and 2 for the definition of these features. The feature importance criteria allow us to evaluate the extent to which each feature contributes to the success of the automated pain assessment model and serves as a mechanism to generate a hypothesis of potential relationship that can be tested directly through experiments.

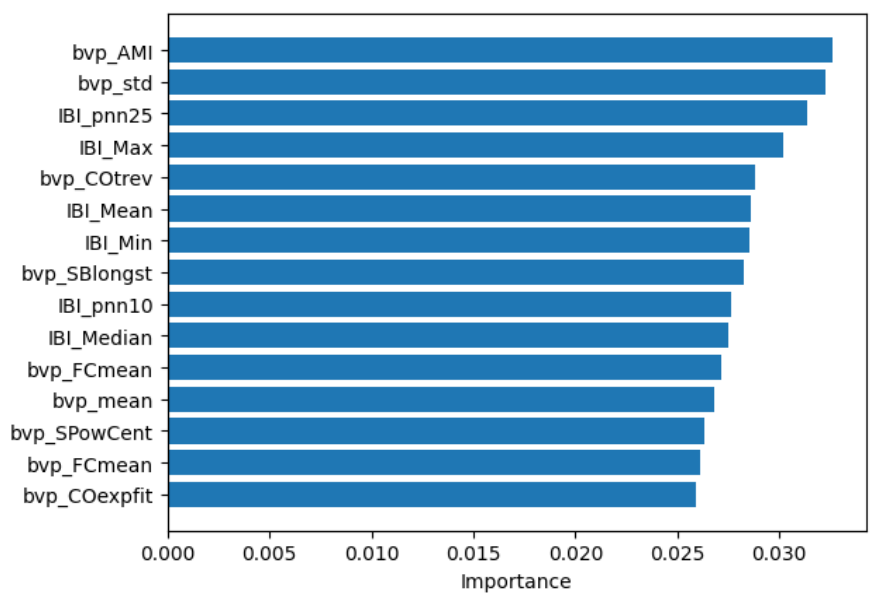

**Figure 6.** Feature importance using ExtraTree estimator on BVP and IBI for pain intensity assessment.

Our results showed that the Auto-Mutual Information (AMI) from the histogram of BVP signal was the most important feature for pain level classification. The auto-mutual information (AMI) function describes the amount of common information between the original time series $x_i$ and the time-shifted time series $x_{i+\tau}$. The AMI function is a nonlinear equivalent of the auto-correlation function based on the Shannon entropy. AMI can be calculated as:

$$AMI(\tau) = \frac{1}{d-1}\log_2 \sum_{x_t \in X}\sum_{x_{t+\tau} \in X} \frac{P_{xx}{}^d(x_t, x_{t+\tau})}{P_x{}^{d-1}(x_t)P_x{}^{d-1}(x_{t+\tau})}$$

where $P(x_t)$ is the estimated probability functions, $d$ is the control parameter, and $\tau$ is the embedding delay. AMI of the signal shows how the signal can predict its time-shifted version. A signal with higher AMI values denotes better predictability. Figure 7 depicts the AMI feature from the histogram of the BVP signal at different pain states. The Dunn's test verified a significant difference in histogram AMI of BVP signal in the presence or absence of pain, in all pain levels from low to medium and high pain. Moreover, this AMI feature from the BVP signal is significantly different in high pain compared to both low pain and medium pian ($p$-value <0.05). It indicates the importance of the auto mutual information as a significant nonlinear biomarker from the BVP signal for pain intensity estimation.

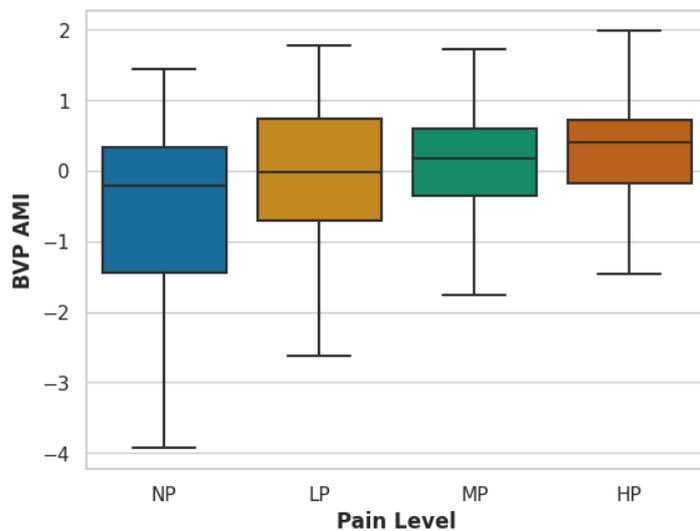

**Figure 7.** Auto Mutual Information (AMI) feature from the histogram of BVP signal at different pain states. There is a significant difference between no pain and each of the other pain levels (*p-value* <0.05). Moreover, the AMI feature of subjects experiencing high pain is significantly different from other pain levels (*p-value* <0.05).

As can be seen in Figure 8, the standard deviation of the BVP signal is a good feature for detecting the presence or absence of cold pain, which shows a significant difference across pain and no pain states (*p-value* <0.05). This feature could not only significantly distinguish the pain and no pain states, but also it could significantly differentiate the high pain from medium pain and low pain (*p-value* <0.05 ). On the other hand, the standard deviation of RR intervals also shows a significant difference across pain and no pain states (*p-value* <0.05).

However, this variable couldn't significantly differentiate pain intensity among low, medium, and high pain levels. This finding shows the value of analyzing the raw BVP signal, not just for the heart rate variability measures from corresponding interbeat intervals.

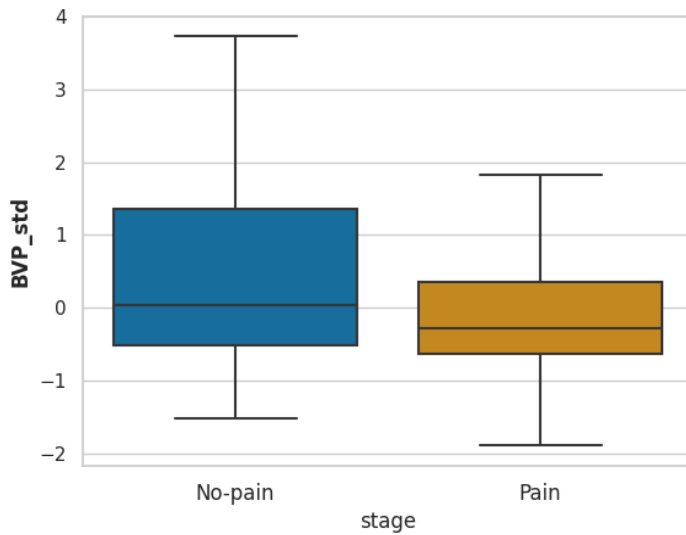

**Figure 8.** Standard deviation BVP signal in presence or absence of pain. This feature has a significant difference (*p-value* <0.05) between pain vs. no pain conditions.

We also explored the features that significantly detected pain, although they couldn't differentiate among the pian levels. One of these features was IBI-sdell which returns the ellipse area fitted into the Poincaré plot of the intervals between heartbeats and computed as SD1*SD2*π. A Poincaré plot, named after Henri Poincaré, is a type of recurrence plot used to quantify self-similarity in processes, usually periodic functions. It is also known as a return map. Poincaré plots can be used to distinguish chaos from randomness by embedding a data set in higher-dimensional state space. SD1 means the standard deviation of the Poincaré plot perpendicular to the line-of-identity and reflects the level of short-term heart rate variability, and SD2 represents the standard deviation of the Poincaré plot along the line-of-identity and indicates the level of long-term heart rate variability[34]. Figure 9 shows the changes in this feature during the cold pressor test. The Poincaré plot has been explored in the literature as a heart rate variability measure in other domains such as stress evaluation[35] or detecting the Atrial Fibrillation[36]. However, to the best of our knowledge, our current work for the first time identifies the potential value of the Poincaré plot of interbeat intervals for automated pain detection. The other exciting feature we found in this experiment was the IBI_Max, which was significantly different when the participants had no or low pain than when they experienced medium or high pain. Figure 10 represents the changes in the maximum interbeat interval during the cold pressor test.

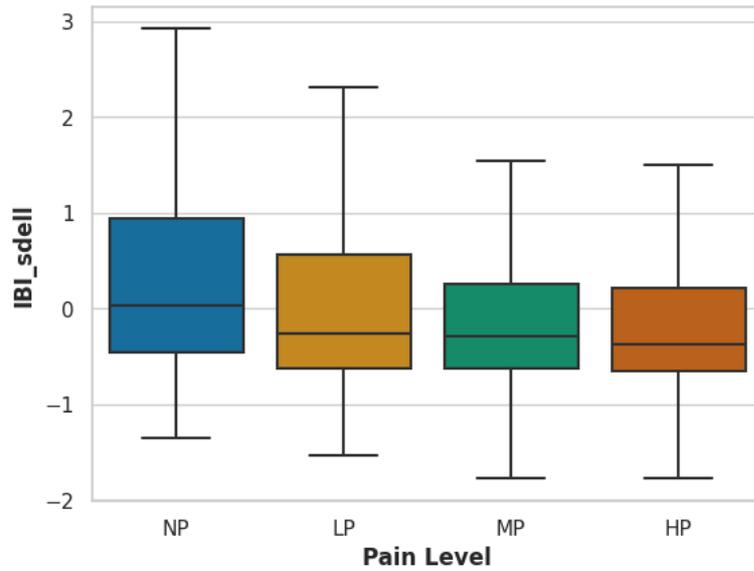

**Figure 9**. The SD1*SD2*π value of the Poincaré plot of input Inter Beat Intervals is significantly different (*p-value* <0.05) between subjects experiencing no pain and each of other pain levels (low, medium, and high pain). Thus, IBI_Sdell is a valuable feature for detecting the presence of pain.

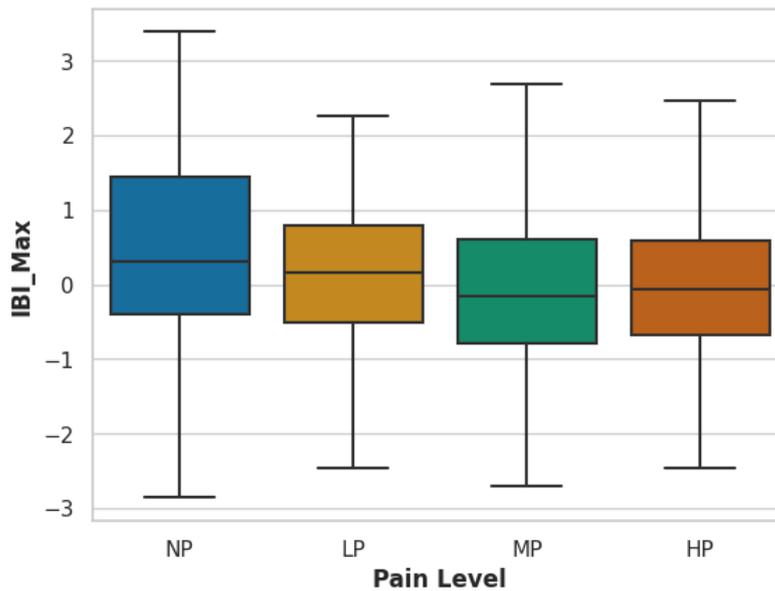

**Figure 10.** There is a significant difference (*p-value* <0.05) between Maximum IBIs of subjects when they are experiencing no or low pain vs. when they are experiencing medium or high pain.

## Conclusion

In this study, we explored BVP signal for automated pain assessment using machine learning algorithms. BVP signals were collected from thirty-two healthy subjects during the cold pressor test. We explored a novel set of 24 features directly from BVP signal, and 20 heart rate variability features from BVP inter-beat intervals from time-domain, frequency-domain, and nonlinear dynamics metrics. We explored different machine learning models: Logistic Regression, SVR, Random Forest, AdaBoost, and XGBoost. In general, among all the models, XGBoost gave the best model performance for both pain classification and pain intensity estimation tasks. The XGBoost pain detection model to detect low pain, medium pain and high pain with no pain as the baseline achieved an average F1-score of $80.06 \pm 5.14\%$, $85.81 \pm 5.06\%$, $90.05 \pm 2.67\%$ respectively. For continuous pain estimation on a numerical rating scale between 1-10, the XGBoost model achieved an average MAE of $0.94 \pm 0.04$ and an average RMSE of $1.26 \pm 0.06$. We also achieved an average F1-score of $80.03 \pm 1.28\,\%$ for multiclass classification among low, medium, and high pain. To the best of our knowledge, this is the first study that investigated this novel set of features from BVP signal for pain detection and pain intensity estimation. Our results show that BVP is a promising non-invasive signal for automated pain assessment.

For future work, we plan to continue our efforts in investigating the performance of other non-invasive signals collected through the cold pressor test for automated pain assessment. We will also explore data fusion techniques to find the possible benefits of combing different modalities. Although we validated our model with healthy subjects and received promising results, it would be valuable if we could also validate with pain patients in a clinical context to investigate the effect of specific diseases on physiological responses to pain.

## Acknowledgments

This material is based upon work supported by the National Science Foundation's Division of Information and Intelligent Systems under Grant No. 1838796. Any opinions, findings, and conclusions or recommendations expressed in this material are those of the authors and do not necessarily reflect the views of the National Science Foundation.

## Author contributions

FP contributed to conceptualization, data curation, formal analysis, investigation, methodology, visualization, writing the original draft, review and editing. YL contributed to funding acquisition, resources, review and editing. SK contributed to funding acquisition, supervision, investigation, resources, validation, project administration, review and editing. All authors reviewed and approved the manuscript.

## Competing interests

The authors declare no competing interests.

## Data availability

All the data used in this study are managed and protected under IRB#: 191215 approved by the Northeastern University human subject research protection Office. We obtained consent from all the participants involved for participation in the experiments. All methods were carried out in accordance with relevant guidelines and regulations.